# Is there a 'Simple' Machine Learning Method for Commonsense Reasoning?

## A Short Commentary on Trinh & Le (2018)

WALID S. SABA

*Astound.ai*
*Menlo Park, CA*

*This is a short Commentary on Trinh & Le (2018) ("A Simple Method for Commonsense Reasoning") that outlines three serious flaws in the cited paper and discusses why data-driven approaches cannot be considered as serious models for the commonsense reasoning needed in natural language understanding in general, and in reference resolution, in particular.*

## 1. INTRODUCTION

In recent years Levesque (2011) and Levesque et. al. (2012) have suggested what they termed the Winograd Schema (WS) challenge as an alternative to the Turing Test for machine intelligence, at least as it relates to human-level language understanding[1]. The following is an example test sentence in the WS challenge:

(1)   *The **trophy** does not fit into **the suitcase** because it is too*
 a. *small*.
 b. *big*.

A program is then asked the question "what was too small" as a follow-up to (1a), and the question "what was too big" as a follow-up to (1b). Levesque *et. al.* call a sentence such as that in (1) "Google proof" since a system that processed a large corpus cannot "learn" how to resolve such references by finding some statistical correlations in the data, as the only difference between (1a) and (1b) are antonyms that are known to co-occur in similar contexts with the same frequency.

In a recent paper Trinh and Le (2018) - henceforth T&L - suggested that they have successfully formulated a 'simple' machine learning method for performing commonsense reasoning, and in particular, the kind of reasoning that would be required in the process of language understanding. In doing so, T&L use the Winograd Schema (WS) challenge as a benchmark. In simple terms, T&L suggest the following method for "learning" how to successfully resolve the reference "it" in sentences such as those in (1): generate two sentences, s₁, which is (1) with "it" replaced by "the trophy" and s₂, which is (1) with "it" replaced by "the suitcase". T&L then compute, against the backdrop of training on a large corpus (trained on the language model LM-1-Billion, CommonCrawl, and SQuAD), the probabilities of s₁ and s₂ appearing in a large corpus. The substitution that turns out to be more probable, is considered to be the more probable referent of "it".

---

[1] The Winograd Schema challenge was named after Terry Winograd, one of the pioneers of AI, who pointed out (Winograd, 1972) the need for using commonsense knowledge in resolving a reference such as 'they' in sentences such as the following:
 *The city councilmen refused the demonstrators a permit because they*
 a. *feared violence*.
 b. *advocated violence*.
While this is not our primary concern here, and while we believe the WS challenge is in fact an improvement on the traditional tests that were susceptible to deception and trickery, Saba (2018a) suggests that the WS challenge could be further improved by generalizing the test to include, in addition to reference resolution, other linguistic phenomena that also require what we might call *thinking* in the process of language comprehension. What Saba (2018) suggests is that most of what we call *thinking* in the process of language comprehension is about 'discovering' the missing text - text that is almost never explicitly stated but is implicitly assumed as *shared* background knowledge.

There are two possibilities to compute such probabilities: one, that T&L termed "full" which involves computing the probability of a replacement on the full sentence, while the other, termed "partial" is the probability of "is too big" (or "is too small"), conditioned on the replacement in the antecedent. That is:

$P_{full}$(*The trophy doesn't fit in the suitcase because the trophy is too big*)
$P_{partial}$(*is too big | The trophy doesn't fit in the suitcase because the trophy*)

T&L report that the "partial" scoring strategy gave better results. This is in fact reasonable, since one is concerned with computing the probability "is too big" appearing, only if (or conditioned on) "the trophy doesn't fit in the suitcase because the trophy" appearing beforehand, while all other situations should not factor in the computation.

Given this strategy, T&L claim that they have outperformed previous state-of-the-art methods (with 70% accuracy), and, of course, all "without using expensive annotated knowledge bases or hand-engineered features". In what follows we hope to show that (*i*) these reported results are misleading, to say the least; (*ii*) that this 'simple' method does not, **and cannot**, take into consideration the slightest changes in the preferences of what "it" refers to when function words are replaced; and (*iii*) this (and any other data-driven) method will not scale into a workable and reasonable solution since in many cases there is nothing to be replaced (and thus the probabilities are *undefined*) as the actual referent is not even in the data (text), but is an object that is implicitly assumed due to our *shared* background knowledge of the world and the way we talk about it in ordinary spoken language. We will finally conclude with some remarks on data-driven approaches in commonsense reasoning and natural language understanding.

## 2. THREE SERIOUS FLAWS IN T&L

Below we will briefly outline the three main flaws regarding the experiment and the subsequent results reported in T&L, and we do so in order of severity, concluding with the third flaw that we believe points to the theoretical impossibility of data-driven and machine learning reasoning in reference resolution.

### 2.1 EXPONENTIAL NUMBER OF COMBINATIONS:
   WORD EMBEDDINGS DO NOT CAPTURE TYPE SIMILARITIES

The first (and least fatal) issue in the experiment reported in T&L is that the experiment completely ignores the number of variations in a sentence such as (1) that will in the end affect the decision as to what "it" refers to. The point here is this: assuming for current purposes that the method reported in T&L does in fact "learn" what "it" in sentences such as (1) refers to with high accuracy, what was learned, however, is specific to that sentence structure and specific lexical items. Since these methods do not admit symbolic structures, including type hierarchies, there is no relation between the sentence in (1), as far as the data-driven system, and the slightly different, but semantically similar sentences below:

(2) The **trophy** does not fit into **the suitcase** because it is too small/big
The **ball** does not fit into **the suitcase** because it is too small/big
The **laptop** does not fit into **the suitcase** because it is too small/big
The **trophy** does not fit into **the bag** because it is too small/big
etc.

The implications of this are that T&L have to answer the questions of what is the number of examples and what is the time required to train such a model to capture all the (40+ million) combinations, all of which imply different preferences as to what "it" refers to. The astute reader who would still like to salvage the machine learning approach might suggest that the system can learn all these combinations by replacing lexical items such as "trophy" and "suitcase" with semantically similar lexical items using word embeddings (and vector similarity). But that, unfortunately, won't do. The reason for this is that word embeddings (such as word2vec or GloVe) do not capture type-supertype relations, but co-occurrence relationships and thus it cannot be assumed that the probability $P$(*The trophy doesn't fit in the suitcase because the trophy is too big*) will be (almost) the same if "trophy", for example, were to be replaced by a word with a strong vector similarity[2].

In the final analysis, and short of admitting an ontological symbolic structure with type-subtype relations, T&L need to give some realistic numbers regarding the number of training examples and time required to capture the millions of combinations that are possible in a sentence such as that in (1).

## 2.2 FUNCTION WORDS: EQUAL PROBABILITIES, DIFFERENT PREFERENCES

A more serious issue with the 'simple' approach to commonsense reasoning reported by T&L is the issue of function words. In particular, consider again the main computation of the following probability, as suggested by T&L:

$P_{partial}$(*is too big* | *The trophy doesn't fit in the suitcase because the trophy*)

We claim that this probability is equivalent (or near equivalent!) to a number of conditional probabilities that differ only in the choice of the relevant function words. Specifically, we claim the following:

$P_{partial}$(*is too big* | *The trophy doesn't fit in the suitcase because the trophy*)
$\cong P_{partial}$(*isn't too big* | *The trophy doesn't fit in the suitcase although the trophy*)
$\cong P_{partial}$(*is too big* | *The trophy does fit in the suitcase because the trophy*)
$\cong P_{partial}$(*is too big* | *The trophy doesn't fit in the suitcase although the trophy*)
$\cong P_{partial}$(*isn't too big* | *The trophy doesn't fit in the suitcase although the trophy*)
etc.

That is, we claim that function words such as "although" and "because", as well as "is" and "isn't" etc. have, in similar contexts, the same probabilities. The implications of this fact are fatal: the probabilities computed are indistinguishable, although these changes imply completely different preferences regarding what "it" refers to, from the standpoint of commonsense. Since data-driven models cannot account for functional words that clearly

---

[2] For example, in GloVe, "winner" is a very similar word to "trophy" yet one can hardly assume that $P$(*The trophy doesn't fit in the suitcase because the trophy is too big*) is similar to $P$(*The winner doesn't fit in the suitcase because the trophy is too big*)

affect the choice of what the most plausible referent is, such models cannot be seriously considered as a model for language understanding.

### 2.3 IMPLICIT BUT MISSING TEXT: PROBABILITIES CANNOT BE COMPUTED

As serious as the above issues are, there is even a more serious issue that T&L do not address and, in our view, no data-driven approach *can* address. Specifically, data-driven approaches, true to their name, can only make generalizations based on the data they process. However, one of the main challenges in language understanding is 'discovering' what we like to call the 'missing text' - text that is never explicitly stated, but is often implicitly assumed as *shared* background knowledge (See Saba, 2018a and Saba 2018b). The implications for this as it pertains to the data-driven and machine learning approach to commonsense reasoning (in, for example, reference resolution) cannot be overstated. Consider, for example, the following:

(3)  *Dave told everyone in school that he wants to be a guitarist, because he thinks **it** is a great sounding instrument.*

Resolving the reference "it" in such sentences cannot be learned by the method suggested by T&L (nor by any other data-driven method, for that matter). In fact, the probabilities that need to be computed here would be meaningless (they are undefined), since what "it" refers to in (3), namely the abstract type *guitar*, is not even mentioned in the text!

You cannot model what is not there, and thus you cannot make any statistical predictions, and consequently your generalizations are flawed. While the examples given were specific to reference resolution, most of the discussion in this section applies to other phenomena in natural language.

## 3. A CONCLUDING REMARK

The data-driven approach in AI has without a doubt gained considerable notoriety in recent years, and there are a multitude of reasons that led to this fact. While the data-driven approach can provide some useful techniques for practical problems that require some level of natural language **processing** (text classification and filtering, search, etc.), extrapolating the relative success of this approach into problems related to commonsense reasoning, the kind that is needed in true language **understanding**, is not only misguided, but may also be harmful, as this might seriously hinder the field, scientifically and technologically.

### REFERENCES

1. Levesque, H. J. 2011, The Winograd Schema Challenge, In *AAAI 2011 Spring Symposium on Logical Formalizations of Commonsense Reasoning*, pp. 63-68
2. Levesque, H., Davis, E. and Morgenstern, L. (2012), The Winograd Schema Challenge, In *Proceedings of the Thirteenth International Conference on Principles of Knowledge Representation and Reasoning*, AAAI Press, pp. 552-561
3. Saba, W. S. (2018a), On the Winograd Schema Challenge: Levels of Language Understanding and the Phenomenon of the Missing Text, draft available at https://arxiv.org/submit/2415300/view
4. Saba W. S. (2018b), Logical Semantics and Commonsense Knowledge: Where Did we Go Wrong, and How to Go Forward, Again, Under Review, draft available at https://arxiv.org/abs/1808.01741
5. Trinh, T. T. and Le Q. V. (2018), A Simple Method for Commonsense Reasoning, available at https://arxiv.org/pdf/1806.02847.pdf
6. Winograd, T. (1972), *Understanding Natural Language*. New York: Academic Press.